\documentclass{article}

\usepackage{times}
\usepackage{subfig}
\usepackage{enumerate}
\usepackage{amsfonts}
\usepackage{amssymb}
\usepackage{amsmath}
\usepackage{color}
\usepackage{tikz}
\usepackage{url} 
\usepackage[procnames]{listings}
\usepackage{color}
\usepackage{todonotes}
\usepackage{caption}
\usepackage{pgfplots}
\usepackage{colortbl}
\usepackage{hyperref}

\graphicspath{{imgs/}}
\newenvironment{camready}{\color{black}}{\color{black}}

\title{Logic Tensor Networks for \\Semantic Image Interpretation}
\author{Ivan Donadello\\ Fondazione Bruno Kessler and \\ University of Trento \\ Trento, Italy \\ \texttt{donadello@fbk.eu}
\and Luciano Serafini \\Fondazione Bruno Kessler \\ Via Sommarive 18, I-38123 \\ Trento, Italy \\ \texttt{serafini@fbk.eu} \and Artur d'Avila Garcez \\ City, University of London \\ Northampton Square \\ London EC1V 0HB, UK \\ \texttt{a.garcez@city.ac.uk}}

\begin{document}
\newenvironment{changed}{\color{black}}{\color{black}}
\newcommand{\argmin}{\mathop{\mathrm{argmin}}}
\newcommand{\argmax}{\mathop{\mathrm{argmax}}}
\newtheorem{theorem}{Theorem}[section]
\newtheorem{lemma}[theorem]{Lemma}
\newtheorem{proposition}[theorem]{Proposition}
\newtheorem{corollary}[theorem]{Corollary}
\newtheorem{definition}{Definition}

\def\PPics{\Pics^{\textsf{PP}}}
\def\PPicst{\Pics^{\textsf{PP}}_t}
\def\PPicse{\Pics^{\textsf{PP}}_e}
\def\gbase{\G_{base}}
\def\tensorflow{\textsc{TensorFlow}$^{TM}$}
\def\AO{\mathit{AO\!}}
\def\pascalpart{\textsc{PASCAL-Part}}
\def\BOmega{\mathbf{\Omega}}
\def\Citation{\mathit{Cite}}
\def\Contains{\mathit{Contains}}
\def\C{\mathcal{C}}
\def\Fn{\mathcal{F}}
\def\G{\mathcal{G}}
\def\Influential{\mathit{Influential}}
\def\KG{\left<\K,\G\right>}
\def\K{\mathcal{K}}
\def\L{\mathcal{L}}
\def\MoreRecent{\mathit{MoreRecent}}
\def\Pics{\mathcal{P}}
\def\Pr{\mathcal{P}}
\def\Rel{\mathcal{R}}
\def\R{\mathbb{R}}
\def\Seminal{\mathit{Seminal}}
\def\Similar{\mathit{Similar}}
\def\T{\mathcal{T}}
\def\animal{\mathsf{animal}}
\def\arity{\alpha}
\def\ba{\mathbf{a}}
\def\bb{\mathbf{b}}
\def\bc{\mathbf{c}}
\def\bt{\mathbf{t}}
\def\bu{\mathbf{u}}
\def\bv{\mathbf{v}}
\def\bv{\mathbf{v}}
\def\bw{\mathbf{w}}
\def\bx{\mathbf{x}}
\def\cat{\mathsf{Cat}}
\def\coach{\mathsf{Coach}}
\def\dog{\mathsf{Dog}}
\def\headof{\mathsf{headOf}}
\def\head{\mathsf{head}}
\def\horse{\mathsf{horse}}
\def\imp{\rightarrow}
\def\leg{\mathsf{leg}}
\def\ltn{\texttt{ltn}}
\def\muzzle{\mathsf{Muzzle}}
\def\partOfPerson{\mathsf{partOfPerson}}
\def\partof{\mathsf{partOf}}
\def\person{\mathsf{person}}
\def\pof{pof}
\def\tail{\mathsf{Tail}}
\def\tailof{\mathsf{tailOf}}
\def\torso{\mathsf{torso}}
\def\train{\mathsf{Train}}
\def\Nat{\mathbb{N}}
\def\bW{\mathbf{W}}
\def\term{\mathit{term}}
\def\pG{\hat{\mathcal{G}}}
\def\GT{\langle\K,\pG\rangle}
\def\vwphi{\langle[v,w],\phi\rangle}
\def\vwphiarg#1{\langle[v,w],\phi(#1)\rangle}
\def\vwphix{\vwphiarg{\bx}}
\def\vwphit{\vwphiarg{\bt}}
\def\bt{\mathbf{t}}
\def\bx{\mathbf{x}}
\def\Loss{\text{Loss}}
\def\parg{\theta}
\def\F{\mathcal{F}}
\def\AI{\mathit{AI}}
\def\GG{\mathbb{G}}
\def\ir{\mathit{ir}}
\def\pics{\mathit{Pics}}
\def\pOmega{\hat{\Omega}}
\def\gwo{\G_{\mathrm{expl}}}
\def\gw{\G_{\mathrm{prior}}}
\def\gfrcnn{\G_{\mathrm{FRCNN}}}
\def\grbpof{\G_{\mathrm{RBPOF}}}
\def\gtwo{\mathcal{T}_{\mathrm{expl}}}
\def\gtw{\mathcal{T}_{\mathrm{prior}}}
\def\kwo{\mathcal{K}_{\mathrm{expl}}}
\def\kw{\mathcal{K}_{\mathrm{prior}}}

\maketitle

\begin{abstract}

  Semantic Image Interpretation (SII) is the task of extracting structured semantic descriptions from images. It is widely agreed that the combined use of visual data and background knowledge is of great importance for SII. Recently, Statistical Relational Learning (SRL) approaches have been developed for reasoning under uncertainty and learning in the presence of data and rich knowledge. Logic Tensor Networks (LTNs) are an SRL framework which integrates neural networks with first-order fuzzy logic to allow (i) efficient learning from noisy data in the presence of logical constraints, and (ii) reasoning with logical formulas describing general properties of the data. In this paper, we develop and apply LTNs to two of the main tasks of SII, namely, the classification of an image's bounding boxes and the detection of the relevant \emph{part-of} relations between objects. To the best of our knowledge, this is the first successful application of SRL to such SII tasks. The proposed approach is evaluated on a standard image processing benchmark. Experiments show that the use of background knowledge in the form of logical constraints can improve the performance of purely data-driven approaches, including the state-of-the-art Fast Region-based Convolutional Neural Networks (Fast R-CNN). Moreover, we show that the use of logical background knowledge adds robustness to the learning system when errors are present in the labels of the training data. \end{abstract}

\section{Introduction}
\emph{Semantic Image Interpretation (SII)} is the task of generating a structured semantic description of the content of an image. This structured description can be represented as a labelled directed graph, where each vertex corresponds to a bounding box of an object in the image, and each edge represents a relation between pairs of objects; verteces are labelled with a set of object types and edges are labelled by the binary relations. Such a graph is also called a \emph{scene graph} in \cite{krishnavisualgenome}.

A major obstacle to be overcome by SII is the so-called \emph{semantic gap} \cite{neumann2008onscene}, that is, the lack of a direct correspondence between low-level features of the image and high-level semantic descriptions. To tackle this problem, a system for SII must learn the latent correlations that may exist between the numerical features that can be observed in an image and the semantic concepts associated with the objects. It is in this learning process that the availability of relational background knowledge can be of great help. Thus, recent SII systems have sought to combine, or even integrate, visual features obtained from data and symbolic knowledge in the form of logical axioms \cite{FeiFei2014Affordance,Chen2012Understanding,donadello2016integration}.

The area of Statistical Relational Learning (SRL), or Statistical Artificial Intelligence (StarAI), seeks to combine data-driven learning, in the presence of uncertainty, with symbolic knowledge \cite{wang2008hybrid,DBLP:journals/corr/BachBHG15,gutmann2010extending,diligenti2015semantic,injecting-knowledge-in-relation-extraction-naacl-2015,learning_in_hybrid_domains_ravkic_2015}. However, only very few SRL systems have been applied to SII tasks (c.f. Section \ref{sec:related-work}) due to the high complexity associated with image learning. Most systems for solving SII tasks have been based, instead, on deep learning and neural network models. These, on the other hand, do not in general offer a well-founded way of learning from data in the presence of relational logical constraints, requiring the neural models to be highly engineered from scratch. 

In this paper, we develop and apply for the first time, the SRL framework called Logic Tensor Networks (LTNs) to computationally challenging SII tasks. LTNs combine learning in deep networks with relational logical constraints \cite{serafini-garces-aixia-2016}. It uses a First-order Logic (FOL) syntax interpreted in the real numbers, which is implemented as a deep tensor network. Logical terms are interpreted as feature vectors in a real-valued $n$-dimensional space. Function symbols are interpreted as real-valued functions, and predicate symbols as fuzzy logic relations. This syntax and semantics, called \emph{real semantics}, allow LTNs to learn efficiently in hybrid domains, where elements are composed of both numerical and relational information. 

We argue, therefore, that LTNs are a good candidate for learning SII because they can express relational knowledge in FOL which serves as constraints on the data-driven learning within tensor networks. Being LTN a logic, it provides a notion of logical consequence, which forms the basis for learning within LTNs, which is defined as \emph{best satisfiability}, c.f. Section \ref{sec:learning-and-reasoning}. Solving the best satisfiability problem amounts to finding the latent correlations that may exist between a relational background knowledge and numerical data attributes. This formulation enables the specification of \emph{learning as reasoning}, a unique characteristic of LTNs, which is seen as highly relevant for SII.

This paper specifies SII within LTNs, evaluating it on two important tasks: (i) the classification of bounding boxes, and (ii) the detection of the \emph{part-of} relation between any two bounding boxes. Both tasks are evaluated using the \pascalpart~ dataset \cite{chen14PascalPart}. It is shown that LTNs improve the performance of the state-of-the-art object classifier Fast R-CNN \cite{girshickICCV15fastrcnn} on the bounding box classification task. LTNs also outperform a rule-based heuristic (which uses the \emph{inclusion ratio} of two bounding boxes) in the detection of \emph{part-of} relations between objects. Finally, LTNs are evaluated on their ability to handle errors, specifically misclassifications of objects and part-of relations. Very large visual recognition datasets now exist which are noisy \cite{DBLP:ReedLASER14}, and it is important for learning systems to become robust to noise. LTNs were trained systematically on progressively noisier datasets, with results on both SII tasks showing that LTN's logical constraints are capable of adding robustness to the system, in the presence of errors in the labels of the training data.

The paper is organized as follows: Section \ref{sec:related-work} contrasts the LTN approach with related work which integrate visual features and background knowledge for SII. Section \ref{sec:real-logic} specifies LTNs in the context of SII. Section \ref{sec:learning-and-reasoning} defines the best satisfiability problem in this context, which enables the use of LTNs for SII. Section \ref{sec:evaluation} describes in detail the comparative evaluations of LTNs on the SII tasks. Section \ref{sec:conclusions} concludes the paper and discusses directions for future work.
\section{Related Work}\label{sec:related-work}
The idea of exploiting logical background knowledge to improve SII tasks dates back to the early days of AI. In what follows, we review the most recent results in the area in comparison with LTNs.

Logic-based approaches have used Description Logics (DL), where the basic components of the scene are all assumed to have been already discovered (e.g. simple object types or spatial relations). Then, with logical reasoning, new facts can be derived in the scene from these basic components \cite{neumann2008onscene,Peraldi2009Formalizing}. Other logic-based approaches have used fuzzy DL to tackle uncertainty in the basic components \cite{Hudelot2008Fuzzy,kompa2009FuzzDL,Hudelot2014FCA}. These approaches have limited themselves to spatial relations or to refining the labels of the objects detected. In \cite{donadello2016integration}, the scene interpretation is created by combining image features with constraints defined using DL, but the method is tailored to the \emph{part-of} relation and cannot be extended easily to account for other relations. LTNs, on the other hand, should be able to handle any semantic relation. In \cite{marszalek-schmid-2007-semantic-hierarchies-for-visual-object-recognition,forestier2013coastal}, a symbolic Knowledge-base is used to improve object detection, but only the \emph{subsumption} relation is explored and it is not possible to inject more complex knowledge using logical axioms.

A second group of approaches seeks to encode background knowledge and visual features within \emph{probabilistic graphical models}. In \cite{FeiFei2014Affordance,nyga2014pr2}, visual features are combined with knowledge gathered from datasets, web resources or annotators, about object labels, properties such as shape, colour and size, and affordances, using Markov Logic Networks (MLNs) \cite{Domingos2006Markov} to predict facts in unseen images. Due to the specific knowledge-base schema adopted, the effectiveness of MLNs in this domain is evaluated only for Horn clauses, although the language of MLNs is more general. As a result, it is not easy to evaluate how the approach may perform with more complex axioms. In \cite{DBLP:journals/corr/BachBHG15}, a probabilistic fuzzy logic is used, but not with real semantics. Clauses are weighted and universally-quantified formulas are instantiated, as done by MLNs. This is different from LTNs where the universally-quantified formulas are computed by using an aggregation operation, which avoids the need for instantiating all variables.

In other related work, \cite{Chen2012Understanding,kulkarni2011baby} encode background knowledge into a generic Conditional Random Field (CRF), where the nodes represent detected objects and the edges represent logical relationships between objects. The task is to find a correct labelling for this graph. In \cite{Chen2012Understanding}, the edges encode logical constraints on a knowledge-base specified in DL. Although these ideas are close in spirit to the approach presented in this paper, they are not formalised as in LTNs, which use a deep tensor network and first-order logic, rather than CRFs or DL. In general, the logical theory behind the  functions to be defined in the CRF is unclear. In \cite{kulkarni2011baby}, potential functions are defined as text priors such as co-occurrence of terms found in the image descriptions of Flickr.

In a final group of approaches, here called \emph{language-priors}, background knowledge is taken from linguistic models \cite{ramanathan2015learning,lu2016visual}. In \cite{ramanathan2015learning}, a neural network is built integrating visual features and a linguistic model to predict semantic relationships between bounding boxes. The linguistic model is a set of rules derived from \textsc{WordNet} \cite{fellbaum98wordnet}, stating which types of semantic relationships occur between a subject and an object. In \cite{lu2016visual}, a similar neural network is proposed for the same task but with a more sophisticated language model, embedding in the same vector space triples of the form \emph{subject-relation-object}, such that semantically similar triples are mapped closely together in the embedding space. In this way, even if no examples exist of some triples in the data, the relations can be inferred from similarity to more frequent triples. A drawback, however, is the possibility of inferring inconsistent triples, such as e.g. \emph{man-eats-chair}, due to the embedding. LTNs avoid this problem with a logic-based approach (in the above example, with an axiom to the effect that chairs are not normally edible). LTNs can also handle exceptions, offering a system capable of dealing with crisp axioms and real-valued data, as specified in what follows. 
\section{Logic Tensor Networks}
\label{sec:real-logic}
Let $\L$ be a first-order logic language, whose signature is composed of three disjoint sets $\C$, $\Fn$ and $\Pr$, denoting constants, functions and predicate symbols, respectively. For any function or predicate symbol $s$, let $\arity(s)$ denote its arity. Logical formulas in $\L$ allow one to specify relational knowledge, e.g. the atomic formula $\partof(o_1,o_2)$, stating that object $o_1$ is a part of object $o_2$, the formulae $\forall xy(\partof(x,y)\imp \neg \partof(y,x))$, stating that the relation $\partof$ is asymmetric, or $\forall x (\cat(x) \imp \exists y(\partof(x,y)\wedge\tail(y)))$, stating that every cat should have a tail. In addition, exceptions are handled by allowing formulas to be interpreted in fuzzy logic, such that in the presence of an example of, say, a tailless cat, the above formula can be interpreted naturally as \emph{normally, every cat has a tail}; this will be exemplified later. 

\noindent \textbf{Semantics of $\L$}: We define the interpretation domain as a subset of $\mathbb{R}^n$, i.e. every object in the domain is associated with a $n$-dimensional vector of real numbers. Intuitively, this $n$-tuple represents $n$ numerical features of an object, e.g. in the case of a person, their name in ASCII, height, weight, social security number, etc. Functions are interpreted as real-valued functions, and predicates are interpreted as fuzzy relations on real vectors. To emphasise the fact that we interpret symbols as real numbers, we use the term \emph{grounding} instead of \emph{interpretation}\footnote{In logic, the term \emph{grounding} indicates the operation of replacing the variables of a term or formula with constants or terms that do not contain other variables. To avoid any confusion, we use the synonym \emph{instantiation} for this purpose. It is worth noting that in LTN, differently from MLNs, the instantiation of every first order formula is not required.} in the following definition of semantics. 
\begin{definition} 
  Let $n\in\Nat$. An $n$-\emph{grounding}, or simply \emph{grounding},
  $\G$ for a FOL $\L$
  is a function defined on the signature of $\L$ 
  satisfying the following conditions:
\begin{enumerate}
\itemsep=-\parsep
\item $\G(c)\in\R^n$ for every constant symbol $c\in\C$;
\item $\G(f)\in \R^{n\cdot \arity(f)}\longrightarrow \R^n$ for every $f\in\Fn$;
\item $\G(P)\in \R^{n\cdot \arity(P)}\longrightarrow [0,1]$ for every
  $P\in\Pr$. 
\end{enumerate}
\end{definition}
Given a grounding $\G$, the semantics of closed terms and atomic formulas
is defined as follows:
\begin{align*} 
\G(f(t_1,\dots,t_m)) & = \G(f)(\G(t_1),\dots,\G(t_m)) \\
\G(P(t_1,\dots,t_m)) & = \G(P)(\G(t_1),\dots,\G(t_m))
\end{align*}

\noindent The semantics for connectives is defined according to fuzzy logic; using for instance the Lukasiewicz t-norm\footnote{Examples of t-norms include Lukasiewicz, product and G\"odel. The Lukasiewicz t-norm is $\mu_{Luk}(x,y)=\max(0,x+y-1)$, product t-norm is $\mu_{Pr}(x,y)=x\cdot y$, and G\"odel t-norm is $\mu_{max}(x,y)=\min(x,y)$. See \cite{bergmann2008introduction} for details.}:
\begin{align*}
\G(\neg\phi) & = 1-\G(\phi) \\
\G(\phi\wedge\psi) & = \max(0,\G(\phi)+\G(\psi) - 1) \\ 
\G(\phi\vee\psi) & = \min(1,\G(\phi)+\G(\psi)) \\ 
\G(\phi\imp\psi) & = \min(1,1-\G(\phi)+\G(\psi)) 
\end{align*}

The LTN semantics for $\forall$ is defined in \cite{serafini-garces-aixia-2016} using the $\min$
operator, that is,
$ \G(\forall x \phi(x))=\min_{t\in \term(\L)}\G(\phi(t)))$, where
$\term(\L)$ is the set of instantiated terms of $\L$. This, 
however, is inadequate for our purposes as it does not tolerate exceptions well (the presence of a single exception to the universally-quantified formulae, such as e.g. a cat without a tail, would falsify the formulae. Instead, our intention in SII is that the more examples there are that satisfy a formulae $\phi(x)$, the higher the truth-value of $\forall x \phi(x)$ should be. To capture this, we use for the semantics of $\forall$ a \emph{mean}-operator, as follows:
$$
\G(\forall x \phi(x)) = \lim_{T\rightarrow\term(\L)}mean_p(\G(\phi(t))
\mid t\in T)
$$
where 
$mean_p(x_1,\dots,x_d) = \left(\frac{1}{d}\sum_{i=1}^d
  x_i^{p}\right)^{\frac{1}{p}}$ 
for $p \in \mathbb{Z}$.
\footnote{The popular mean operators, arithmetic, geometric and harmonic mean, are obtained by setting $p=1, 2,$ and $-1$, respectively.} 

Finally, the classical semantics of $\exists$ is uniquely determined
by the semantics of $\forall$, by making $\exists$ equivalent to
$\neg\forall\neg$. This approach, however, has a drawback too when it comes to SII: if we adopt, for instance, the arithmetic mean for the semantic of
$\forall$ then $\G(\forall x \phi(x))=\G(\exists x\phi(x))$. Therefore, we shall interpret existential quantification via Skolemization: every formula of the form
$\forall x_1,\dots,x_n(\dots \exists y \phi(x_1,\dots,x_n,y))$ is
rewritten as
$\forall x_1,\dots,x_n(\dots \phi(x_1,\dots,x_n,f(x_1,\dots,x_n)))$,
by introducing a new $n$-ary function symbol, called Skolem
function. In this way, existential quantifiers can be eliminated from the
language by introducing Skolem functions. 

\noindent \textbf{Formalizing SII in LTNs}:\label{ex:sii}
To specify the SII problem, as defined in the introduction, we consider a
signature $\Sigma_{\text{SII}}=\left<\C,\Fn,\Pr\right>$, where
$\C=\bigcup_{p\in\pics}b(p)$ is the set of identifiers for all the bounding
boxes in all the images, $\Fn=\emptyset$, and 
$\Pr=\{\Pr_1,\Pr_2\}$, where $\Pr_1$ is a set of unary predicates, one for
each object type, e.g. $\Pr_1 = \{\dog, \cat,$
$\tail,\muzzle,\train,\coach,\dots\}$, and $\Pr_2$ is a set of binary
predicates representing relations between objects. Since in our experiments we focus on
the \emph{part-of} relation, $\Pr_2 = \{\partof\}$. The FOL formulas based on this signature can specify (i) simple facts, e.g. the fact that bounding box $b$ contains a cat, written $\cat(b)$, the fact that $b$ contains either a cat or a dog, written $\cat(b)\vee\dog(b)$, etc., and (ii) general rules such as $\forall x (\cat(x)\imp \exists y(\partof(x,y)\wedge \tail(y)))$.

A grounding for $\Sigma_{\text{SII}}$ can be defined as follows: 
each constant $b$, denoting a bounding box, can be associated with 
a set of geometric features and a set of semantic features obtained 
from the output of a bounding box detector. Specifically, each bounding box
is associated with \emph{geometric features} describing the position and the dimension of the bounding box, and \emph{semantic features} describing the classification score returned by the bounding box detector for each class. For example, for each bounding box $b\in\C$, $C_i \in \Pr_1$, $\G(b)$ is the
$\R^{4+|\Pr_1|}$ vector:
  $$
     \langle
     class(C_1,b),\dots,class(C_{|\Pr_1|},b),x_0(b),y_0(b),x_1(b),y_1(b)
     \rangle
  $$
where the last four elements are the coordinates of the top-left and bottom-right corners of $b$, and $class(C_i,b)\in[0,1]$ is the classification score of the bounding box detector for $b$. 

An example of groundings for predicates can be defined by taking a one-vs-all multi-classifier approach, as follows. First, define the following grounding for each class $C_i\in\Pr_1$ (below, $\bx=\left<x_1,\dots,x_{|\Pr_1|+4}\right>$ is the vector corresponding to the grounding of a bounding box):
\begin{align}\label{eq:grBun}
\G(C_i)(\bx)=\left\{
\begin{array}{ll}
1 & \mbox{if } i = \argmax_{1 \leq l \leq |\Pr_1|}x_l \\ 
0 & \mbox{otherwise} 
\end{array}
\right. 
\end{align}
Then, a simple rule-based approach for defining a grounding for the $\partof$ relation is based on the na\"{i}ve assumption that the more a bounding box $b$ is contained within a bounding box $b'$, the higher the probability should be that $b$ is part of $b'$. Accordingly, one can define $\G(\partof(b,b'))$ as the inclusion ratio $\ir(b,b')$ of bounding box $b$, with grounding $\bx$, into bounding box $b'$, with grounding $\bx'$ (formally, $\ir(b,b') = \frac{area(b \cap b')}{area(b)}$). A slightly more sophisticated rule-based grounding for $\partof$ (used as baseline in the experiments to follow) takes into account also \emph{type compatibilities} by multiplying the inclusion ratio by a factor $w_{ij}$. Hence, we define $\G(\partof(b,b'))$ as follows:
    \begin{align}\label{eq:grBpof}
    \left\{
      \begin{array}{ll}
        1 & \mbox{if }  \ir(b,b')\cdot\max_{ij=1}^{|\Pr_1|}(w_{ij}\cdot
                                x_i\cdot x'_j) \geq th_{ir} \\ 
        0 & \mbox{otherwise} 
      \end{array}
      \right.
    \end{align}
for some threshold $th_{ir}$ (we use $th_{ir} > 0.5$), and with $w_{ij}=1$ if $C_i$ is a part of $C_j$, and $0$ otherwise. Given the above grounding, we can compute the grounding of any atomic formula, e.g. $\cat(b_1)$, $\dog(b_2)$, $\leg(b_3)$, $\partof(b_3,b_1)$, $\partof(b_3,b_2)$, thus expressing the degree of truth of the formula. 
The rule-based groundings (Eqs. \eqref{eq:grBun} and \eqref{eq:grBpof}) may not satisfy some of the constraints to be imposed. For example, the classification score may be wrong, a bounding box may include another which is not in the part-of relation, etc. Furthermore, in many situations, it is not possible to define a grounding a priori. Instead, groundings may need to be learned automatically from examples, by optimizing the truth-values of the formulas in the background knowledge. This is discussed next.

\section{Learning as Best Satisfiability}\label{sec:learning-and-reasoning}
A partial grounding, denoted by $\pG$, is a grounding that is defined
on a subset of the signature of $\L$. A grounding $\G$ is said to be
a completion of $\pG$, if $\G$ is a grounding for $\L$ and coincides
with $\pG$ on the symbols where $\pG$ is defined.

\begin{definition}
  A \emph{grounded theory GT} is a pair $\GT$ with a set $\K$ of 
  closed formulas and a partial grounding $\pG$. 
\end{definition}
\begin{definition}
  A grounding $\G$ satisfies a GT $\GT$ if $\G$ completes $\pG$ and
  $\G(\phi)=1$ for all $\phi\in\K$. 
  A GT $\GT$ is \emph{satisfiable} if there exists a grounding $\G$
  that satisfies $\GT$. 
\end{definition}

According to the previous definition, deciding the satisfiability of
$\GT$ amounts to searching for a grounding $\pG$ such that
\emph{all} the formulas of $\K$ are mapped to 1. Differently from the
classical satisfiability, when a GT is
not satisfiable, we are interested in the best possible satisfaction
that we can reach with a grounding. This is defined as
follows.

\begin{definition}
  Let $\GT$ be a grounded theory.  We define the \emph{best
    satisfiability problem} as the problem of finding a grounding
  $\G^*$ that maximizes the truth-values of 
  the conjunction of all clauses $cl \in \K$, i.e. 
  $
  \G^* = \argmax_{\pG\subseteq \G \in \GG}\G(\bigwedge_{cl \in \K}cl).
  $
\end{definition}
Grounding $\G^*$ captures the latent correlation between the quantitative
attribute of objects and their categorical and relational properties. Not all  functions are suitable as a grounding; they
should preserve some form of regularity. If $\G(\cat)(\bx)\approx 1$ (the bounding box with feature vector $\bx$ contains a cat) then for every $\bx'$ close to $\bx$ (i.e. for every bounding box with features similar to $\bx$), one should have $\G(\cat)(\bx')\approx 1$. In particular, we consider groundings of the following form:
 
Function symbols are grounded to linear transformations. If $f$ is a $m$-ary function symbol, then $\G(f)$ is of the form: 
$$
\G(f)(\bv)= M_f\bv + N_f
$$
where $\bv=\left<\bv_1^\intercal,\dots,\bv_m^\intercal\right>^\intercal$ is the $mn$-ary vector obtained by concatenating each $\bv_i$. \begin{camready}The parameters for $\G(f)$ are the $n\times mn$ real matrix $M_f$ and the $n$-vector $N_f$.\end{camready}

The grounding of an $m$-ary predicate $P$, namely $\G(P)$, is defined as a generalization of the neural tensor network (which has been shown effective at knowledge completion in the presence of simple logical constraints \cite{SocherChenManningNg2013}), as a function from $\R^{mn}$ to $[0,1]$, as follows:
\begin{align}\label{eq:grP}
\G(P)(\bv) = \sigma\left(u^\intercal_P\tanh\left(\bv^\intercal W_P^{[1:k]}\bv + V_P\bv + b_P\right)\right)
\end{align}
with $\sigma$ the sigmoid function. \begin{camready}The parameters for $P$ are: $W_P^{[1:k]}$, a 3-D tensor in $\R^{k \times mn\times mn}$, $V_P \in \R^{k\times mn}$, $b_P \in \R^{k}$ and $u_P \in \R^{k}$. This last parameter performs a linear combination of the quadratic features given by the tensor product.\end{camready} With this encoding, the grounding (i.e. truth-value) of a clause can be determined by a neural network which first computes the grounding of the literals contained in the clause, and then combines them using the specific t-norm.

In what follows, we describe how a suitable GT can be built for SII. Let $\pics^t\subseteq\pics$ be a set of bounding boxes of images correctly labelled with the classes that they belong to, and let each pair of bounding boxes be correctly labelled with the part-of relation. In machine learning terminology, $\pics^t$ is a \emph{training set} without noise. In real semantics, a training set can be represented by a theory $\gtwo = \langle\kwo,\pG\rangle$, where $\kwo$ contains the set of closed literals $C_i(b)$ (resp. $\neg C_i(b)$) and $\partof(b,b')$ (resp. $\neg\partof(b,b')$), for every bounding box $b$ labelled (resp. not labelled) with $C_i$ and for every pair of bounding boxes $\left<b,b'\right>$ connected (resp$\neg\partof(b,b')$. not connected) by the $\partof$
  relation. The partial grounding $\pG$ is defined on all bounding
  boxes of all the images in $\pics$ where both the semantic
  features $class(C_i,b)$ and the bounding box coordinates are
  computed by the Fast R-CNN object detector
  \cite{girshickICCV15fastrcnn}. $\pG$ is not defined for the
  predicate symbols in $\Pr$ and is to be learned. $\gtwo$ contains
  only assertional information about specific bounding boxes. This is
  the classical setting of machine learning where classifiers (i.e. the
  grounding of predicates) are inductively learned from positive examples (such as $\partof(b,b')$) and
  negative examples ($\neg\partof(b,b')$) of a classification. In this learning setting, mereological
  constraints such as ``cats have no wheels'' or ``a tail is a part of a cat'' are not taken into account. Examples of
  mereological constraints state, for instance, that the
  part-of relation is asymmetric ($\forall xy(\partof(x,y)\imp\neg\partof(y,x))$), or lists the several parts
  of an object (e.g.
  $\forall xy(\cat(x)\wedge\partof(x,y)\imp\tail(y)\vee\muzzle(y)$)), or even, for simplicity, that every whole object cannot be part of another object (e.g.
  $\forall xy(\cat(x)\imp \neg \partof(x,y)))$ and every part object
  cannot be divided further into parts (e.g. $\forall xy(\tail(x)\imp
  \neg \partof(y,x)))$.
  This general knowledge is available from on-line resources, such as
  \textsc{WordNet} \cite{fellbaum98wordnet}, and can be retrieved by inheriting the meronymy relations for every concept correponding to a whole object. A grounded theory that considers also mereological
  constraints as prior knowledge can be constructed by adding such axioms to $\kwo$. More formally, we define $\gtw=\langle\kw,\pG\rangle$, where $\kw=\kwo+\mathcal{M}$, and $\mathcal{M}$ is the set of mereological axioms. 
  To check the role of $\mathcal{M}$, we evaluate both
  theories and then compare results. 
\section{Experimental Evaluation}\label{sec:evaluation}
\begin{figure*}[t]
  \centering
    \begin{tabular}{cc}  
  
    \subfloat[LTNs with prior knowledge improves the performance of the Fast R-CNN on object type classification, achieving an Area Under the Curve (AUC) of 0.800 in comparison with 0.756.]
   {
   \includegraphics[trim=25 9 32 32,clip,width=.45\linewidth]{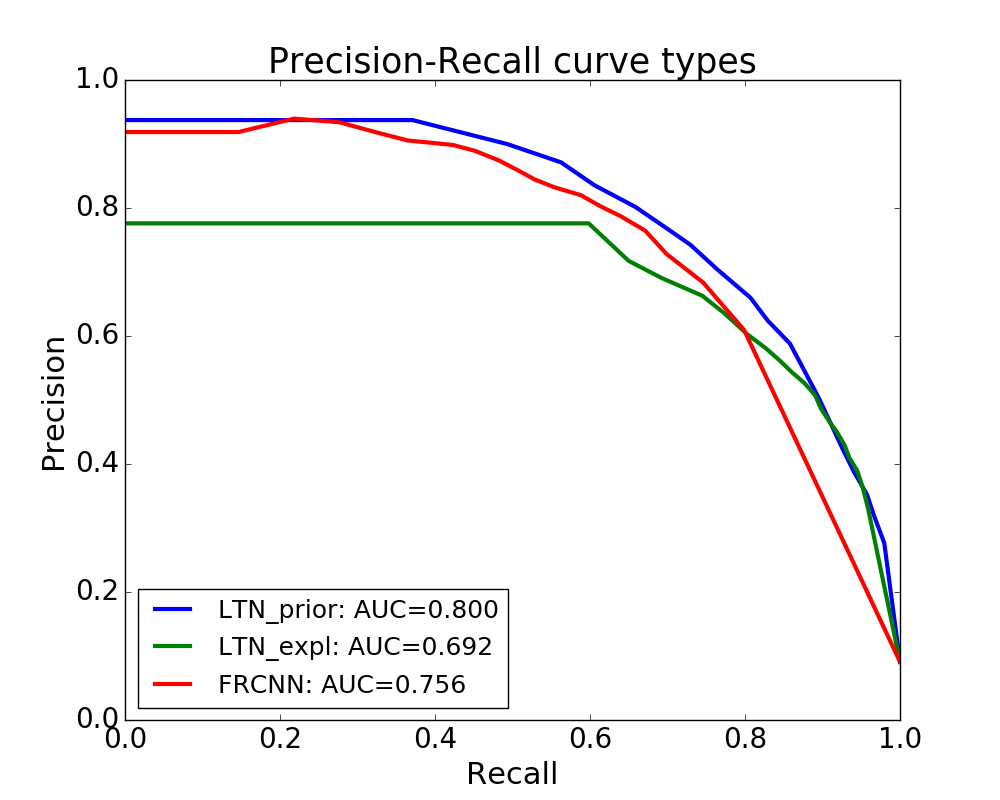}
\label{fig:prec_rec_types}
}
&
\subfloat [LTNs with prior knowledge outperform the rule-based approach of Eq.\ref{eq:grBpof} in the detection of part-of relations, achieving AUC of 0.598 in comparison with 0.172.]
{
\includegraphics[trim=25 9 32 32,clip,width=.45\linewidth]{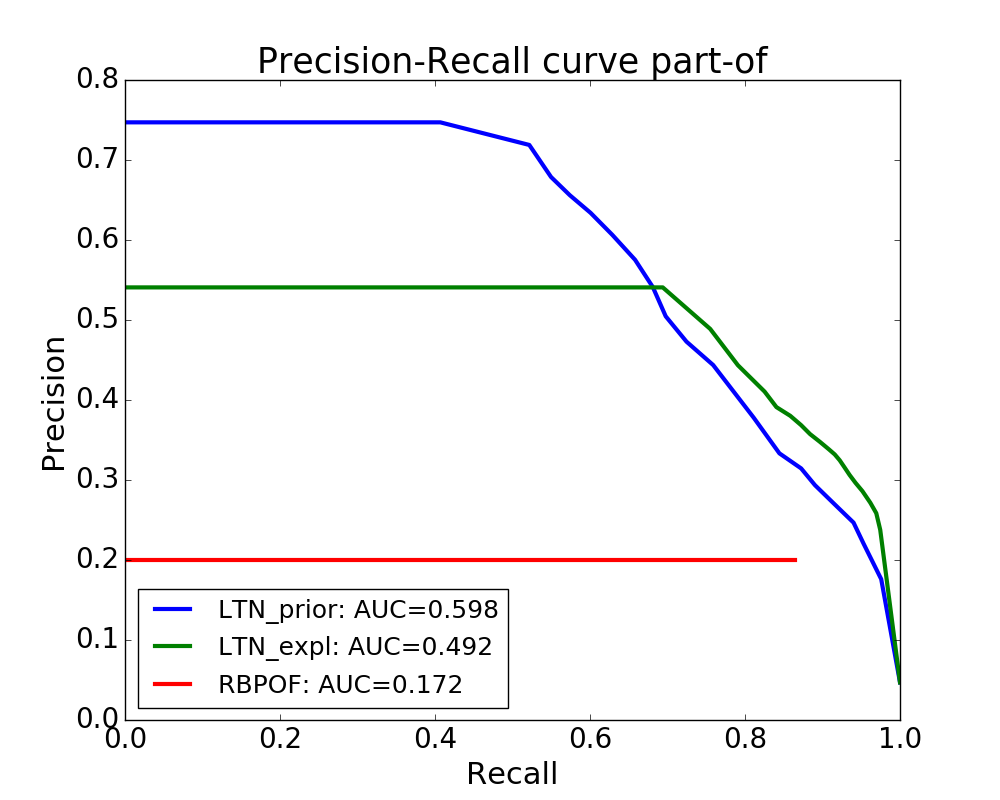} \label{fig:prec_rec_pof}
}
    \end{tabular}

\caption{Precision-recall curves for indoor objects type classification and the $\partof$ relation between objects.}
    \label{fig:prec_rec_curve}
\end{figure*}
We evaluate the performance of our approach for SII\footnote{LTN has been implemented as a Google \tensorflow library. Code, $\partof$ ontology, and dataset are available at \url{https://gitlab.fbk.eu/donadello/LTN_IJCAI17}} on two tasks, namely, the classification of bounding boxes and the detection of $\partof$ relations between pairs of bounding boxes. In particular, we chose the part-of relation because both data (the \pascalpart-dataset \cite{chen14PascalPart}) and ontologies (\textsc{WordNet}) are available on the part-of relation. In addition, part-of can be used to represent, via reification, a large class of relations \cite{DBLP:conf/sebd/GuarinoG16} (e.g., the relation ``a plant is lying on the table'' can be reified in an object of type ``lying event'' whose parts are the plant and the table). \begin{camready}However, it is worth noting that many other relations could have been included in this evaluation. The time complexity of LTN grows linearly with the number of axioms.\end{camready}

We also evaluate the robustness of our approach with respect to noisy data. It has been acknowledged by many that, with the vast growth in size of the training sets for visual recognition \cite{krishnavisualgenome}, many data annotations may be affected by noise such as missing or erroneous labels, non-localised objects, and disagreements between annotations, e.g. human annotators often mistake ``part-of'' for the ``have'' relation \cite{DBLP:ReedLASER14}. 

We use the \pascalpart-dataset that contains 10103 images with bounding boxes annotated with object-types and the part-of relation defined between pairs of bounding boxes. Labels are divided into three main groups: animals, vehicles and indoor objects, with their corresponding parts and ``part-of'' label. Whole objects inside the same group can share parts. Whole objects of different groups do not share any parts. Labels for parts are very specific, e.g. ``left lower leg''. Thus, without loss of generality, we have merged the bounding boxes that referred to the same part into a single bounding box, e.g. bounding boxes labelled with ``left lower leg'' and ``left upper leg'' were merged into a single bounding box of type ``leg''. In this way, we have limited our experiments to a dataset with 20 labels for whole objects and 39 labels for parts. In addition, we have removed from the dataset any bounding boxes with height or width smaller than 6 pixels. The images were then split into a training set with 80\%, and a test set with 20\% of the images, maintaining the same proportion of the number of bounding boxes for each label.

\textbf{Object Type Classification and Detection of the Part-Of Relation}: Given a set of bounding boxes detected by an object detector (we use Fast-RCNN), the task of object classification is to assign to each bounding box an object type. The task of Part-Of detection is to decide, given two bounding boxes, if the object contained in the first is a part of the object contained in the second. We use LTN to resolve both tasks simultaneously. This is important because a bounding box type and the part-of relation are not independent. Their dependencies are specified in LTN using background knowledge in the form of logical axioms. 

To show the effect of the logical axioms, we train two LTNs: the first containing only training examples of object types and part-of relations ($\gtwo$), and the second containing also logical axioms about types and part-of ($\gtw$). The LTNs were set up with tensor of $k=6$ layers and a regularization parameter $\lambda=10^{-10}$. We chose Lukasiewicz's T-norm ($\mu(a,b) = \max(0,a+b-1)$) and use the harmonic mean as aggregation operator. We ran 1000 training epochs of the RMSProp learning algorithm available in \tensorflow. 
We compare results with the Fast RCNN at object type classification (Eq.\eqref{eq:grBun}), and the \emph{inclusion ratio} $ir$ baseline (Eq.{eq:grBpof}) at the part-of detection task\footnote{\begin{camready} A direct comparison with \cite{Chen2012Understanding} is not possible because their code was not available.\end{camready}}. If $ir$ is larger than a given threshold $th$ (in our experiments, $th$=0.7) then the bounding boxes are said to be in the $\partof$ relation. \begin{camready}Every bounding box $b$ is classified into $C \in \Pr_1$ if $\G(C(b))\geq th$. With this, a bounding box can be classified into more than one class. For each class, precision and recall are calculated in the usual way.\end{camready} 
Results for indoor objects are shown in Figure \ref{fig:prec_rec_curve} where AUC is the area under the precision-recall curve. The results show that, for both object types and the part-of relation, the LTN trained with prior knowledge given by mereological axioms has better performance than the LTN trained with examples only. Moreover, prior knowledge allows LTN to improve the performance of the Fast R-CNN (FRCNN) object detector. Notice that the LTN is trained using the Fast R-CNN results as features. \begin{camready}FRCNN assigns a bounding box to a class if the values of the corresponding semantic features exceed $th$. This is local to the specific semantic features. If such local features are very discriminative (which is the case in our experiments) then very good levels of precision can be achieved. Differently from FRCNN, LTNs make a global choice which takes into consideration all (semantic and geometric) features together. This should offer robustness to the LTN classifier at the price of a drop in precision. The logical axioms compensate this drop.\end{camready} For the other object types (animals and vehicles), LTN has results comparable to FRCNN: FRCNN beats $\gtw$ by 0.05 and 0.037 AUC, respectively, for animals and vehicles. Finally, we have performed an initial experiment on \emph{small data}, on the assumption that the LTN axioms should be able to compensate a reduction in training data. By removing 50\% of the training data for indoor objects, a similar performance to $\gtw$ with the full training set can be achieved: 0.767 AUC for object types and 0.623 AUC for the part-of relation, which shows an improvement in performance.  

\begin{figure}
    \centering
        
    \subfloat[Object types]{%
        \includegraphics[trim=25 9 32 32,clip,width=0.6\linewidth]{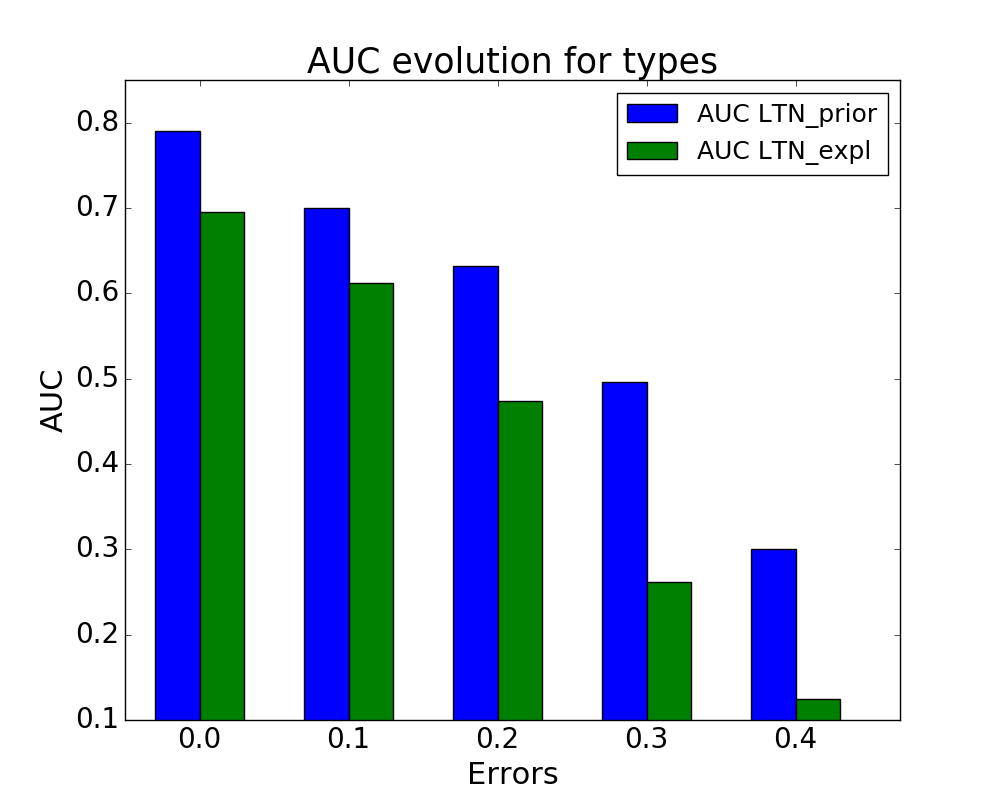}%
        \label{fig:AP_part-types}%
        }%
    
    \subfloat[Part-of predicate]{%
        \includegraphics[trim=25 9 32 32,clip,width=0.6\linewidth]{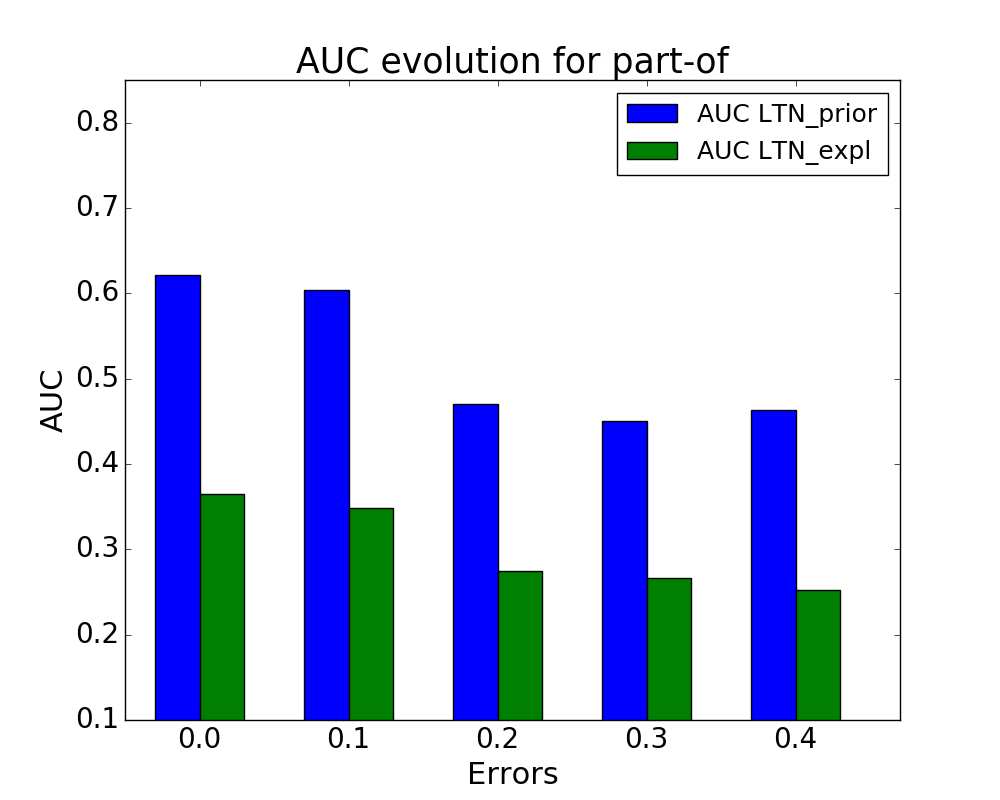}%
        \label{fig:AP_part-of}%
        }%
    
    \caption{AUCs for indoor object types and part-of relation with increasing noise in the labels of the training data. The drop in performance is noticiably smaller for the LTN trained with background knowledge.}
    \label{fig:recovery_noisy_labels}
\end{figure}

\textbf{Robustness to Noisy Training Data}: In this evaluation, we show that logical axioms improve the robustness of LTNs in the presence of errors in the labels of the training data. We have added an increasing amount of noise to the \pascalpart-dataset training data, and measured how performance degrades in the presence and absence of axioms. For $k\in\{10,20,30,40\}$, we randomly select $k\%$ of the bounding boxes in the training data, and randomly change their classification labels. In addition, we randomly select $k\%$ of pairs of bounding boxes, and flip the value of the part-of relation's label. For each value of $k$, we train LTNs $\gtwo^k$ and $\gtw^k$ and evaluate results on both SII tasks as done before. As expected, adding too much noise to training labels leads to a large drop in performance. Figure \ref{fig:recovery_noisy_labels} shows the AUC measures for indoor objects with increasing error $k$. Each pair of bars indicates the AUC of $\gtw^k, \gtwo^k$, for a given $k\%$ of errors. 
Results indicate that the LTN axioms offer robustness to noise: in addition to the expected overall drop in performance, an increasing gap can be seen between the drop in performance of the LTN trained with exampels only and the LTN trained including background knowledge.

\section{Conclusion and Future Work}\label{sec:conclusions}
SII systems are required to address the semantic gap problem: combining visual low-level features with high-level concepts. We argue that the problem can be addressed by the integration of numerical and logical representations in deep learning. LTNs learn from numerical data and logical constraints, enabling approximate reasoning on unseen data to predict new facts. In this paper, LTNs were shown to improve on state-of-the-art method Fast R-CNN for bounding box classification, and to outperform a rule-based method at learning part-of relations in the \pascalpart-dataset. Moreover, LTNs were evaluated on how to handle noisy data through the systematic creation of training sets with errors in the labels. Results indicate that relational knowledge can add robustness to neural systems. As future work, we shall apply LTNs to larger datasets such as \textsc{Visual Genome}, and continue to compare the various instances of LTN with SRL, deep learning and other neural-symbolic approaches on such challenging visual intelligence tasks.
\clearpage
\bibliographystyle{plain}
\bibliography{bibliography}

\end{document}